\documentclass[runningheads]{llncs}

\usepackage{eccv}

\usepackage{eccvabbrv}

\usepackage[utf8]{inputenc} %
\usepackage[T1]{fontenc}    %
\usepackage{graphicx}
\usepackage{xspace}
\usepackage{amsmath}
\usepackage{amssymb}
\usepackage{pifont}
\usepackage{booktabs}
\usepackage{multicol,multirow}
\usepackage{changes}
\usepackage{xcolor}         %
\usepackage{url}            %
\usepackage{array}       %
\usepackage{colortbl}
\usepackage{algorithm}
\usepackage{bm}
\usepackage[algo2e]{algorithm2e}
\usepackage{amsfonts}       %
\usepackage{nicefrac}       %
\usepackage{microtype}      %
\usepackage{mathtools}
\usepackage{alias}
\usepackage{lib}
\usepackage{cuted}
\usepackage[labelfont=bf]{caption}
\usepackage{wrapfig}
\usepackage{subcaption}
\usepackage{enumitem}

\usepackage[accsupp]{axessibility}  %

\usepackage{hyperref}

\usepackage{orcidlink}

\begin{document}

\title{Bimanual 3D Hand Motion and Articulation Forecasting in Everyday Images}

\author{Aditya Prakash\inst{1} \quad Richard Li\inst{2} \quad David Forsyth\inst{1} \quad Saurabh Gupta\inst{1}}

\authorrunning{A.~Prakash et al.}

\institute{University of Illinois Urbana-Champaign \and Massachusetts Institute of Technology\\
\url{https://bit.ly/ForeHand4D}}

\maketitle

\begin{abstract}
    We tackle the problem of forecasting bimanual 3D hand motion and articulation from a single image in everyday settings. To address the lack of 3D hand annotations in diverse settings, we design an annotation pipeline consisting of a diffusion model to lift 2D hand keypoint sequences to 4D hand motion. For the forecasting model, we adopt a diffusion loss to account for the multimodality in hand motion distribution. Extensive experiments on 6 datasets show the benefits of training with our imputed labels (14\% improvement) and the effectiveness of our lifting (45\% better) \& forecasting (16.4\% gain) models, over the best baselines, especially in zero-shot generalization to everyday images.
\end{abstract}

\section{Introduction}
\label{sec:intro}
\begin{figure*}
    \centering
    \setlength{\tabcolsep}{0pt}
    \begin{tabular}{c}
        \toprule
        \includegraphics[width=\textwidth]{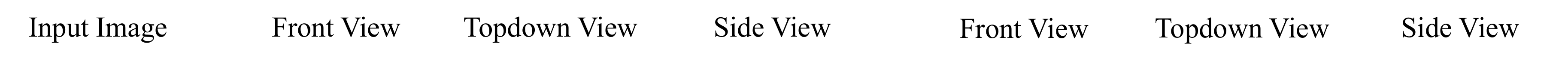} \\ [-0.5em]
        \midrule 
        \includegraphics[width=\textwidth]{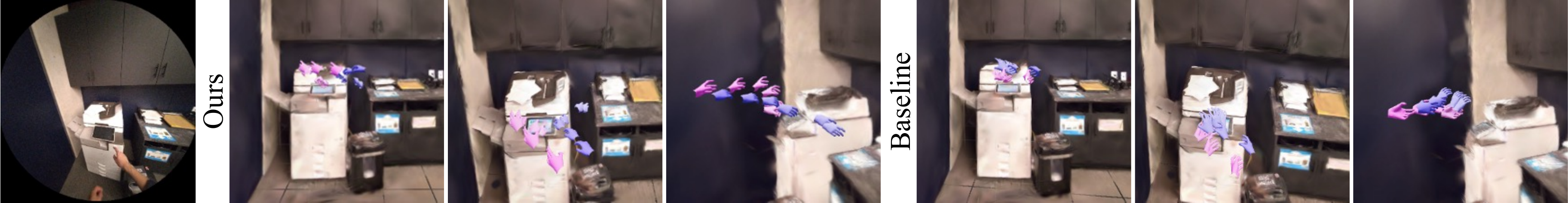} \\ [-0.25em]
        \midrule
        \includegraphics[width=\textwidth]{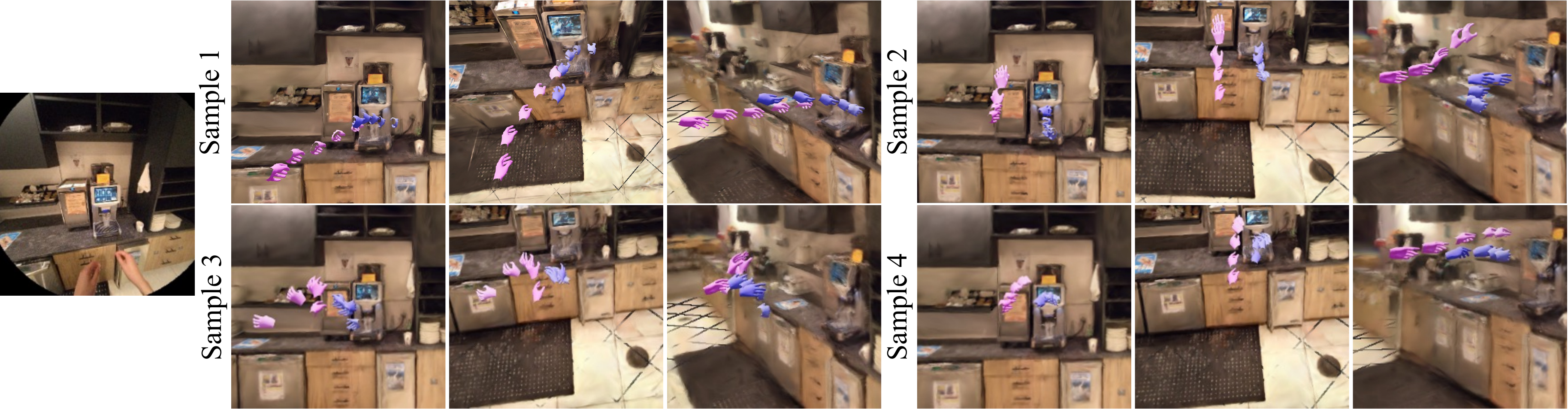} \\ [-0.25em]
        \bottomrule
    \end{tabular}
    \vspace{-0.2cm}
    \caption{
    \name forecasts bimanual 3D hand motion from a single RGB image input. (top) The predicted motions (shown from 3 views) highlight object interactions, \eg reaching for the paper on the table beside the printer, compared to the baseline where the hand is in air. (bottom) \name can also generate multiple motions from the same input, \eg reaching for the coffee machine/cabinet, leftward grabbing action or pointing to the screen.  Left hand shown in {\color{magenta}pink}, right hand in {\color{blue}blue}. Color saturation decreases as time proceeds, \ie, further out timesteps are denoted by lighter shades. The 3D scene is only for visualization \& these images are not seen during training.}
    \label{fig:teaser}
    \vspace{-0.5cm}
\end{figure*}

This paper develops \name, a system for {\it forecasting bimanual 3D hand
motion} from a single everyday RGB image as input. \name can operate on {\it
diverse everyday images} to generate the {\it full articulation} of the hand in
{\it 3D} for {\it both} hands over {\it long} time horizons while only requiring {\it a single RGB image}. This expands capability along several axes: generalization, prediction horizon \& completeness of output; thereby improving the utility of such models for downstream human-robot interaction, AR/VR coaching or animation applications. \Eg, accurate forecasts for what a human's hand might do next can prevent a co-operating robot from colliding with it \& better yet offer more meaningful assistance. Or an AR coaching app can show how a human should modify their grasp of a tennis racket. Or generated motions can offer suggestions for an animator to pick from. \cref{fig:teaser} shows sample outputs from \name, including on images from EgoExo4D not used for training in any way. Generations from \name are more coherent than the baseline ($1^{\text{st}}$ row) and different samples capture different possible interactions from the same starting point ($2^{\text{nd}}$ row).

Generating hand motion is difficult because of the complex ways in which hands
interact with one another and the surrounding environment.  Because hands can
do many different things in the future (\ie output is multi-modal), training a
regressor is not suitable. Therefore, we adopt a diffusion loss for
training our models.
We find that this successfully mitigates problems arising due to multi modality
and leads to a large improvement in forecasting performance on a suite of lab
datasets as shown in \cref{tab:transformer}, representing the first
experimental results on this challenging problem.

However, there is a mismatch between the data we can train such a diffusion models on (lab datasets where \textit{complete} 3D ground truth, \ie MANO~\cite{romero2017embodied} parameters, are available) \vs the data we would like this diffusion model to work on (everyday images outside of lab settings, that may have some 2D annotations but no 3D labels). Because a diffusion model needs complete ground truth for training (the forward diffusion process adds noise to the ground truth before denoising), prior techniques~\cite{tulsiani2017multi, yan2016perspective,pavlakos2024reconstructing} that leverage weak supervision via reprojection losses are not applicable since MANO parameters are not available. Generating 3D pseudo-labels from available 2D annotations is the obvious solution but existing methods, \eg EasyMocap~\cite{easymocap}, that directly optimize the MANO parameters using 2D reprojection loss, are not effective since they are highly sensitive to initialization, optimization objective \& hyper-parameter tuning. Our innovation is to develop a {\it learned lifting model} that lifts available 2D annotations into complete 3D annotations. This increases the diversity of data for training the forecasting model, and thereby its performance on held-out datasets (\cref{tab:transformer}).

\begin{figure*}[t]
\centering
\insertW{1.0}{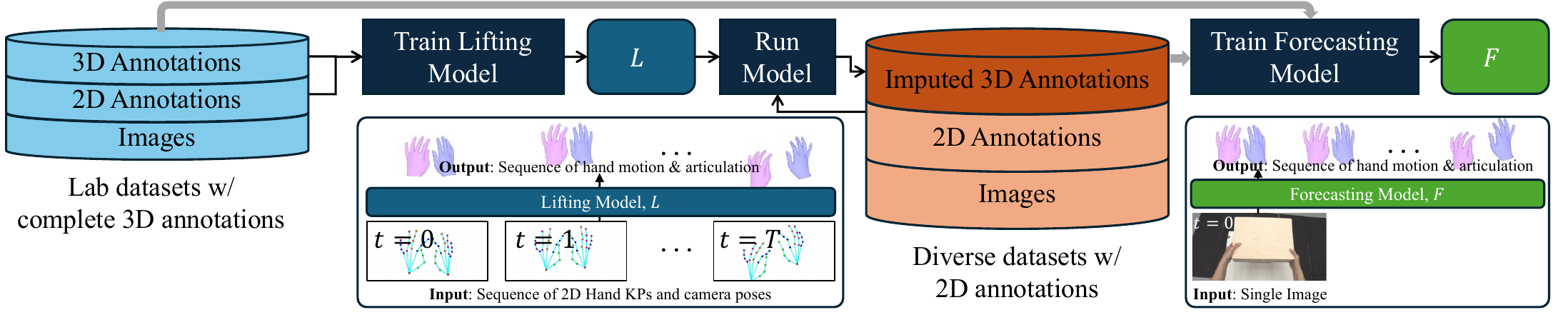}
\vspace{-0.7cm}
\caption{\textbf{Overall Training Pipeline.}
We first use the 2D \& 3D annotations in lab datasets to train a lifting
diffusion model, $\liftingD$ that maps 2D keypoints sequences to 3D MANO hands. We
then run $\liftingD$ on diverse datasets with 2D annotations to generate 3D
annotations. Finally, the forecasting model $\foreD$
is trained on lab \& diverse datasets with complete 3D supervision.}
\label{fig:meta-arch}
\vspace{-0.5cm}
\end{figure*}

Our overall pipeline is shown in \cref{fig:meta-arch}. We first use 3D annotated
lab datasets to develop the lifting diffusion model, $\liftingD$. This model
takes as input 2D hand keypoints \& camera parameters (intrinsics \&
extrinsics) across all timesteps in a sequence to output the corresponding 3D hand articulation \& placement. We use $\liftingD$ to lift 2D annotations on diverse datasets into 3D ground truth. We then use these imputed 3D labels, alongside true 3D ground truth labels on lab datasets to train our forecasting diffusion model, $\foreD$. Because 2D annotations are more readily available on diverse datasets, this increases the data diversity for training the forecasting model. 
Experiments reveal that training on diverse data, enabled by our method, substantially improves the
predictions of learned models on both everyday images from EgoExo4D (zero-shot generalization) \& lab datasets (\cref{tab:transformer}), and outperforms
LatentAct~\cite{prakash2025how} (adapted \& retrained to work in our
setting) by 16.4\%. It also improves over alternative ways of injecting weak
supervision via auxiliary 2D forecasting heads. Also, our lifting
model generates more accurate 3D labels, 65.3\% better than the recent HaWoR method~\cite{Zhang2025ARXIV} (\cref{tab:assembly}). Code \& models are available at \url{https://github.com/ap229997/forehand4d}.

\section{Related Work}
\label{sec:related}

\textbf{3D Hand Prediction}. Given image or video input,
many recent works make \textit{predictions} (and not forecasts) for hands
present in them from egocentric~\cite{Fan2023CVPR,Fan2024ECCV,li2023ego,kwon2021h2o,Ye2026haptic,Wen2023HTT} and exocentric observations~\cite{hampali2020honnotate, Hampali2022CVPR,Moon2020ECCV,
yi2024estimating, rong2020frankmocap, potamias2024wilor, Lin2021CVPR, Moon2022CVPRW, Zhang2023PAMI,ohkawa2023assemblyhands}. HaMeR~\cite{pavlakos2024reconstructing} is a high-performing recent work that predicts 3D hands from single images, while Dyn-HaMR~\cite{yu2024dyn} makes temporally consistent 3D hand
predictions from videos. Predictions on videos are made via feed-forward model~\cite{ye2025predicting}, test-time optimization~\cite{Zhang2025ARXIV} or hybrid methods~\cite{Hewitt2024TOG}. 

\noindent \textbf{4D Hand Forecasting}. Prior works have looked at
forecasting specific aspects of hand motion in different settings. Works differ
in what they output and from what input. On the output side: \cite{Cha2024CVPR,
prakash2025how} produce hand motion, articulation \& contact maps, \cite{Hatano2025EgoH4,bao2023uncertainty} forecast the 3D wrist location, while Liu
\etal~\cite{liu2022joint} forecast only the 2D wrist location. On the input
side, \cite{liu2022joint, bao2023uncertainty, yang20223d,
cao2021reconstructing,Gavryushin2025ARXIV} only use RGB images as input, while
\cite{Cha2024CVPR, zhang2024bimart, zhang2024manidext, prakash2025how, christen2024diffh2o} are conditioned on privileged information in the form of articulating 3D objects or 3D contact points on objects. Papers also tackle different settings: full body forecasting~\cite{yang20223d, cao2021reconstructing}, single hand-object interaction~\cite{prakash2025how, bao2023uncertainty,Gavryushin2025ARXIV,zhou2025megohand} or bimanual interactions~\cite{Cha2024CVPR, liu2022joint,zhang2024manidext, zhang2024bimart, christen2024diffh2o}. Thus, past work addresses individual aspects of the problem, but none as comprehensively as ours: they either produce rich 3D output from stronger input (3D object models) or use RGB images but predict only coarse 2D/3D results.

\begin{figure*}
\centering
\insertW{1.0}{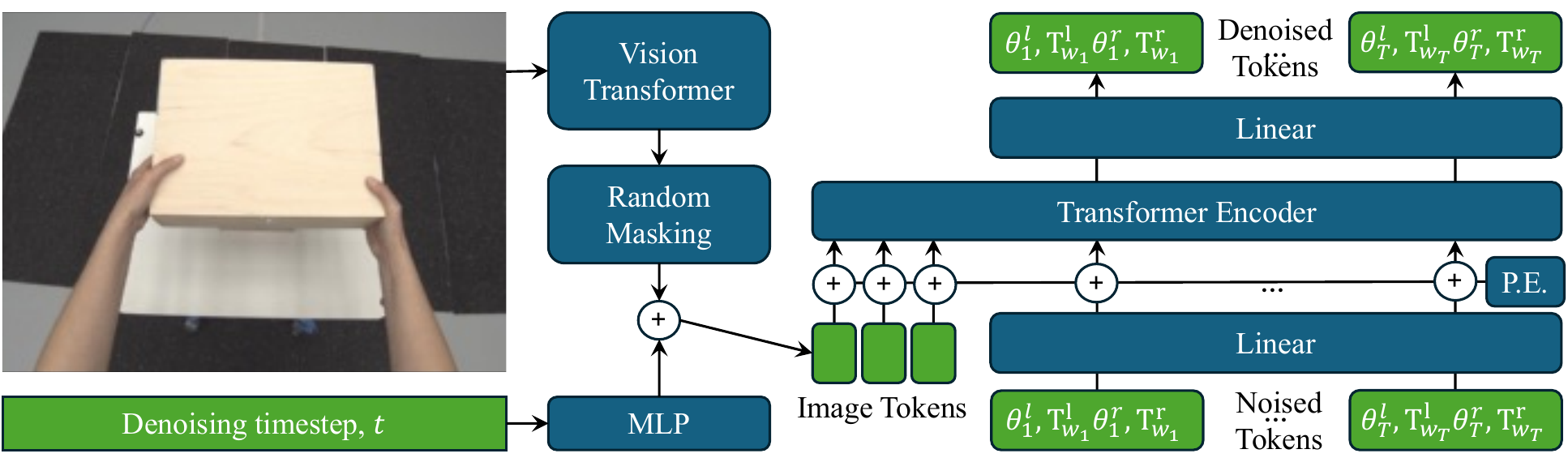}
\vspace{-0.8cm}
\caption{\textbf{Architecture for Forecasting Model}. We modify MDM~\cite{Tevet2023ICLR} to condition on images features extracted from a ViT backbone. Each input \& output token is 198-dimensional: 2 hands $\times$ (16 (joints) $\times$ (6 (6D rotation for each joint) + 3 (wrist translation))). 
}
\label{fig:forecasting-arch}
\vspace{-0.6cm}
\end{figure*}

\noindent \textbf{3D pose from 2D keypoints}. Several works in the human pose literature have explored estimating 3D pose from 2D keypoints using different approaches, \eg linear models~\cite{Ramakrishna2012ECCV}, probabilistic models~\cite{Simo-Serra2013CVPR}, directly optimizing 3D poses~\cite{Akhter2015CVPR}, MLP~\cite{Martinez2017ICCV,Moreno-Noguer2017CVPR}, convolutional~\cite{Pavallo2019CVPR,Moreno-Noguer2017CVPR,Tome2017CVPR}, graph-based~\cite{Cai2019ICCV,Ci2019ICCV,Wang2020ECCV}, transformer~\cite{Li2022CVPR,Shan2022ECCV,Zhang2022CVPR,Zheng2021ICCV}, diffusion~\cite{Li2025CVPR,Kapon2024CVPR} \& normalizing flows~\cite{Wandt2022CVPR}. These works operate on different types of inputs, \eg static 2D pose~\cite{Akhter2015CVPR,Martinez2017ICCV,Wandt2022CVPR}, 2D pose estimated from image~\cite{Moreno-Noguer2017CVPR,Simo-Serra2013CVPR}, sequence of 2D keypoints~\cite{Cai2019ICCV,Wang2020ECCV,Shan2022ECCV,Zhang2022CVPR,Zheng2021ICCV,Kapon2024CVPR} or multi-view 2D poses~\cite{Li2025CVPR}.
These ideas have also been extended to estimated 3D hand poses using 2D keypoints~\cite{Zimmermann2017ICCV,Xu2025HandBooster}, 2D/2.5D heatmaps~\cite{Hampali2022CVPR,Iqbal2018ECCV} or depth~\cite{Cheng2024HandDiff}. Building on top of these works, we design a diffusion-based lifting model to estimate 3D hand poses from 2D keypoints to scale up 3D hand annotations for diverse settings.

\noindent \textbf{Learning from Incomplete 3D Ground Truth}. Prior works often
inject weak supervision into 3D regression models via a 2D reprojection loss~\cite{tulsiani2017multi, yan2016perspective,pavlakos2024reconstructing}:
the predicted 3D is differentiably rendered or projected into 2D and encouraged
to match the 2D annotations.
However, this doesn't naturally extend to diffusion models that need to know the score
$\nabla_\mathbf{x} \log p(\mathbf{x})$ at different
locations $\mathbf{x}$. While we can render a denoised 3D shape and compute
partial supervision on it using the reprojection loss (\ie we don't know
$\nabla_\mathbf{x} \log p(\mathbf{x})$ but only a projection of it), we don't know
what point $\mathbf{x}$ in space is this the partial supervision for. 
Recent works explore training diffusion models on corrupted or partial data~\cite{bai2024expectation,Daras2023NEURIPS} using EM~\cite{bai2024expectation, hastie2009elements} or aggressive masking~\cite{Daras2023NEURIPS}. Our setting is different because we don't quite have partial ground truth, but rather a projection of a 3D shape into 2D.

\section{4D Hand Forecasting}
\label{sec:method}

Given a single RGB image $I$ showing a hand object scenario, the task is to forecast the 3D hand motion for both hands. We use MANO hand representation~\cite{Romero2017TOG}, consisting of the shape $\beta$, articulation $\theta$ \& global wrist pose $\wrist{}{c}{}{w}$, where $c$ is the world frame located at the camera center. The goal is to learn a
function $\foreD(I)$ that takes the image $I$ as input and predicts $\Phi_{t} =
\{(\theta^l_t, \wrist{}{c}{l}{w_t}, \theta^r_t, \wrist{}{c}{r}{w_t})\}$ for all timesteps in the prediction horizon, where $l$ \& $r$ superscripts denote left \& right hand. We do not predict $\beta$ (we use the mean $\beta$ shape from the MANO model when $\beta$ is not available). 

\begin{figure*}
\centering
\insertW{1.0}{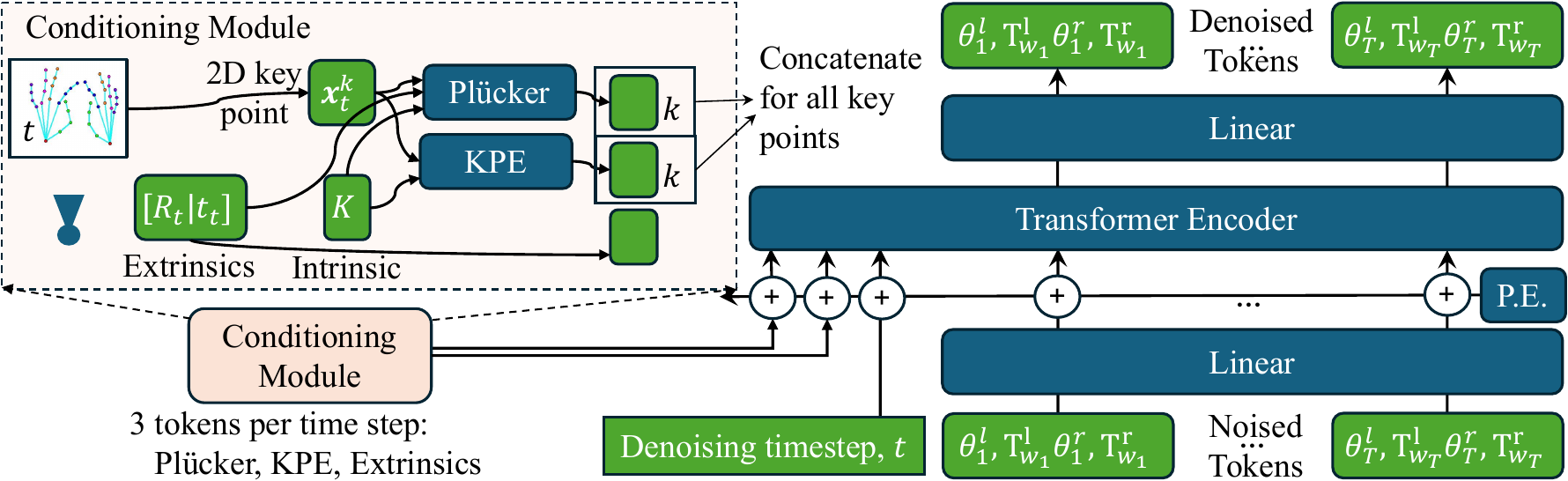}
\vspace{-0.5cm}
\caption{\textbf{Architecture for Lifting Model}. We modify MDM~\cite{Tevet2023ICLR} to condition on a sequence of 2D hand keypoints \& camera parameters. The conditioning module combines different input representations: 3D pose (rotation, translation) of camera, Pl\"ucker rays~\cite{Zhang2024ICLR} \& KPE~\cite{Prakash2023Ambiguity}.}
\label{fig:lifting-arch}
\vspace{-0.5cm}
\end{figure*}

Our forecasting model $\foreD$ is realized using a transformer \& trained
using a diffusion loss (\cref{sec:arch}). A diffusion loss means we need {\it complete} 3D
annotations for training. Thus, we can only use carefully
constructed lab datasets with complete 3D annotations (\eg ARCTIC~\cite{Fan2023CVPR}, H2O~\cite{kwon2021h2o}, H2O-3D~\cite{Hampali2022CVPR}, HOT3D~\cite{Banerjee2024ARXIV}, \& DexYCB~\cite{Chao2021CVPR}) for training $\foreD$. This severely limits the diversity
of data that $\foreD$ is exposed to, and thereby its generalization
capabilities. To mitigate this limitation, we develop a {\it lifting model,}
$\liftingD$ to lift 2D key point annotations to complete 3D annotations
(\cref{sec:lifting}). Our final forecasting model is trained on the union of 3D
lab datasets, and 2D in-the-wild datasets lifted to 3D, as shown in 
\cref{fig:meta-arch}.

\subsection{Forecasting Model, $\foreD$}
\label{sec:arch}
Since temporal forecasts are multimodal, we adopt a conditional diffusion
model to represent $\foreD$, \ie given the inputs \& noisy versions of the desired
outputs, $\foreD$ predicts the noise that was added to the outputs. $\foreD$ uses a ViT~\cite{Dosovitskiy2021ICLR}
backbone to encode the image $I$. We modify the diffusion model from
MDM~\cite{Tevet2023ICLR} for our setting. Specifically, we change the
conditioning to provide image features as input.
Following~\cite{Tevet2023ICLR}, we use a transformer encoder for the denoising (\cref{fig:forecasting-arch}) and 6D
representation~\cite{Zhou2019CVPR} for rotation. All the 3D poses are represented in the camera coordinate frame at $t=0$ and the predictions are also done in the camera frame at $t=0$.

\subsection{Lifting Model, $\liftingD$}
\label{sec:lifting}
The lifting model takes as input 2D hand keypoints \& camera parameters over a
sequence to output the 3D hand placement (\ie wrist translation and rotation) \& articulation (\ie joint angles) in MANO representation in the camera frame from the first frame. It is realized using a conditional diffusion model with a
transformer backbone (\cref{fig:lifting-arch}). 

{\bf Conditioning Module.} Because 2D hand keypoints and the camera parameters are intertwined, we concatenate different representations to use as conditioning to the diffusion model: {\bf (1) Extrinsics}: 6D rotation representation and 3D translation. {\bf (2) Pl\"ucker rays}: These encode the camera rays joining the camera
center with the 2D keypoints in the image. Specifically, let $\bm{x}^k_t = (x^k_t, y^k_t, 1)$
denote the 2D location of the $k^\text{th}$ hand keypoint at time
step $t$ in homogeneous coordinates, $\bm{K}$ denote the camera intrinsic parameters,
and $\bm{R}_t, \bm{t}_t$ denote the camera rotation and translation, such that
$\bm{K}[\bm{R}_t | \bm{t}_t] \bm{X}$ maps a world point $\bm{X}$ into the camera
frame. The ray joining the camera center to $\bm{x}^k_t$ is given by
$\lambda \bm{R}^{-1}_t \bm{K}^{-1}\bm{x}^k_t - \bm{K}^{-1}\bm{t}_t$. We represent this ray
using the 6D Pl\"ucker representation~\cite{Zhang2024ICLR}. {\bf (3) KPE~\cite{Prakash2023Ambiguity}}: This captures the location of each 2D keypoint in the field of view of the camera (with principal point $(p_x, p_y)$ and focal length $(f_x, f_y)$). For each $(x^k_t, y^k_t)$, we estimate the angles $\phi_x =
\tan^{-1} ((x^k_t - p_x)/{f_x})$ and $\phi_y = \tan^{-1} (({y^k_t - p_y})/{f_y})$ and compute sinusoidal encodings.

{\bf Training Data.} We render out 3D hand
and camera trajectories in the lab datasets into 2D hand keypoints trajectories.
We also introduce augmentations in the camera trajectories to increase diversity
in data for training. Because of these augmentations and not using any visual
information, the lifting model generalizes very well to datasets not seen during
training.

\subsection{Using Lifting Model, $\liftingD$, to Impute 3D Labels}
\label{sec:training}
We impute MANO labels by running the lifting model $\liftingD$ on 2D annotations in diverse datasets (AssemblyHands~\cite{ohkawa2023assemblyhands} \& HoloAssist~\cite{Wang2023ICCV}). Rather than
directly using the 3D output from the lifting model, we adjust the 3D output to
get it to better conform to the 2D annotations. Concretely, we pass the complete 3D predictions from the lifting model to the differentiable MANO
model, $\mathcal{M}$, to get the 3D hand joints, which are then projected into
the image with known intrinsics to get 2D keypoints. We optimize the reprojection loss on 2D keypoints labels (either available in datasets like
AssemblyHands or estimated from off-the-shelf model~\cite{pavlakos2024reconstructing}) using gradient descent for 1000 iterations with a learning rate of 0.01 \& gradient norm clipped to 1 for regularization. For datasets with 3D keypoint labels (but no MANO labels), we also add a L2 loss on 3D keypoints.

\subsection{Implementation Details}
\label{sec:implementation}

The denoiser in both $\foreD$ \& $\liftingD$ is implemented as a transformer encoder with 16 layers, 4 heads, latent dimension of 1024 \& dropout of 0.1. The ViT backbone for computing image features is initialized from~\cite{pavlakos2024reconstructing}. Following~\cite{Tevet2023ICLR}, we use 1000 steps for denoising with cosine noise schedule. We also mask out the conditioning tokens (image features for $\foreD$ and 2D keypoints + camera parameters for $\liftingD$) with probability 0.1 to simulate noise in diverse settings. For augmentation, we add pixel-level noise \& image scaling for forecasting and we jitter \& scale 2D keypoints for lifting. Both $\foreD$ \& $\liftingD$ predict normalized translation value (using mean \& standard deviation across all the wrist translations in the training dataset). For $\liftingD$, we find the combination of extrinsics, plucker rays \& KPE to work the best. The predictions span 256 timesteps.
Both $\foreD$ \& $\liftingD$ are trained in a mixed-dataset setting across 4 NVIDIA L40S or 4 A40 GPUs. Since different datasets have varying length sequences, we mask out the extra timesteps.

\begin{table*}
\centering
\caption{\textbf{Datasets}. We train jointly on all the datasets using the available MANO labels and incomplete supervision from HoloAssist (2D keypoint labels are from off-the-shelf HaMeR~\cite{pavlakos2024reconstructing}) \& AssemblyHands. Different datasets have varying number of objects, sequences \& time horizon (in seconds) and we train a single model over all these variable factors. Note that EgoExo4D is only used for testing the \emph{zero-shot generalization} performance (not for training), and the \emph{lifting model is not trained on AssemblyHands}.}
\label{tab:datasets}
\vspace{-0.2cm}
\resizebox{1.0\linewidth}{!}
{
\begin{tabular}{l cccc ccccc}
\toprule
\bf Dataset Name                      & \bf View & \bf Lab/Wild & \bf Labels & \bf \#Motions & \bf \#Objs & \bf Horizon(s) & \bf Role ($\liftingD$) & \bf Role ($\foreD$)\\
\midrule
ARCTIC~\cite{Fan2023CVPR}     & Ego       & Lab        & MANO    &   4499  & 11               & 0.5-5 & train & train (MANO), test \\
H2O~\cite{kwon2021h2o}            & Ego       & Lab        & MANO    &  534   &  8            & 1-7.27 & train & train (MANO), test \\
H2O3D~\cite{Hampali2022CVPR}            & Exo       & Lab        & MANO    &  57   & 10       & 0.8-4.87 & train & train (MANO) \\
HOT3D~\cite{Banerjee2024ARXIV}    & Ego       & Lab       & MANO    &  6000   &  33           & 5 & train & train (MANO), test \\
DexYCB~\cite{Chao2021CVPR}        & Exo       & Lab        & MANO    &  5743   &  20          & 1.77-1.97 & train & train (MANO), test \\
\midrule
HoloAssist~\cite{Wang2023ICCV}    & Ego       & Wild       & 2D Kps  &  7461   &  120         & 0.53-8.5 & --  & train ($\liftingD(\text{2D Kps}$) \\
AssemblyHands~\cite{ohkawa2023assemblyhands} 
                                  & Ego       & Lab        & 3D + 2D Kps    &  2134   & 101   & 0.53-8.5 & test & train ($\liftingD(\text{2D + 3D Kps})$), test \\
\midrule
EgoExo4D~\cite{Grauman2023EgoExo4D}  & Ego & Wild       & 3D Kps   &  53   &  --             & 0.53-6.57 & -- & zero-shot testing \\
\bottomrule
\end{tabular}}
\vspace{-0.5cm}
\end{table*}

\section{Experiments}
\label{sec:expts}
We focus on (a) how does our lifting approach compare to other ways of injecting 2D supervision, (b) the quality of our imputed labels, (c) how well does our model forecast hand motion from a single image \wrt related past methods, (d) the effectiveness of a diffusion head
over a regression head, (e) does incorporating diverse 2D labeled data help?

\noindent \textbf{Metrics}. 
We adopt metrics from human motion
literature~\cite{Yuan2022CVPR,Shin2024CVPR,Ye2023CVPRa,Guo2024CVPR,Tevet2023ICLR,guo2020action2motion} that measure the accuracy and quality of predicted motions. For accuracy, we use: (a)
Mean Per Joint Position Error (\ma) in 3D (in cm), averaged over time, keypoints
\& 2 hands, (b) Mean Relative-Root Position Error (\md) in 3D (in cm) that
measures the translation between the root joint of left \& right hand. We also
include two variants of \ma: (c) \mb (predictions are \textit{globally aligned} to
the ground truth before computing \ma) \& (d) \mc (predictions are aligned to the
ground truth at the \textit{first timestep}). By doing some form of alignment,
\mb \& \mc focus on the accuracy of the predicted articulation. Lower is better. 

For quality (forecasting task only): (a) Diversity: measures the variance over a set of motions across the full dataset of predicted or ground truth motions. We represent motions by stacking MANO parameters over time and compute the mean pairwise L2 distance among the motions. As in~\cite{guo2020action2motion}, the predicted diversity should be comparable to that of the ground truth motions. (b) Multimodality: measures the variation within forecasted motions from the same input, computed as the mean pairwise L2 distance between 5 samples per input (higher is better).

\begin{figure*}
    \centering
    \includegraphics[width=0.495\textwidth]{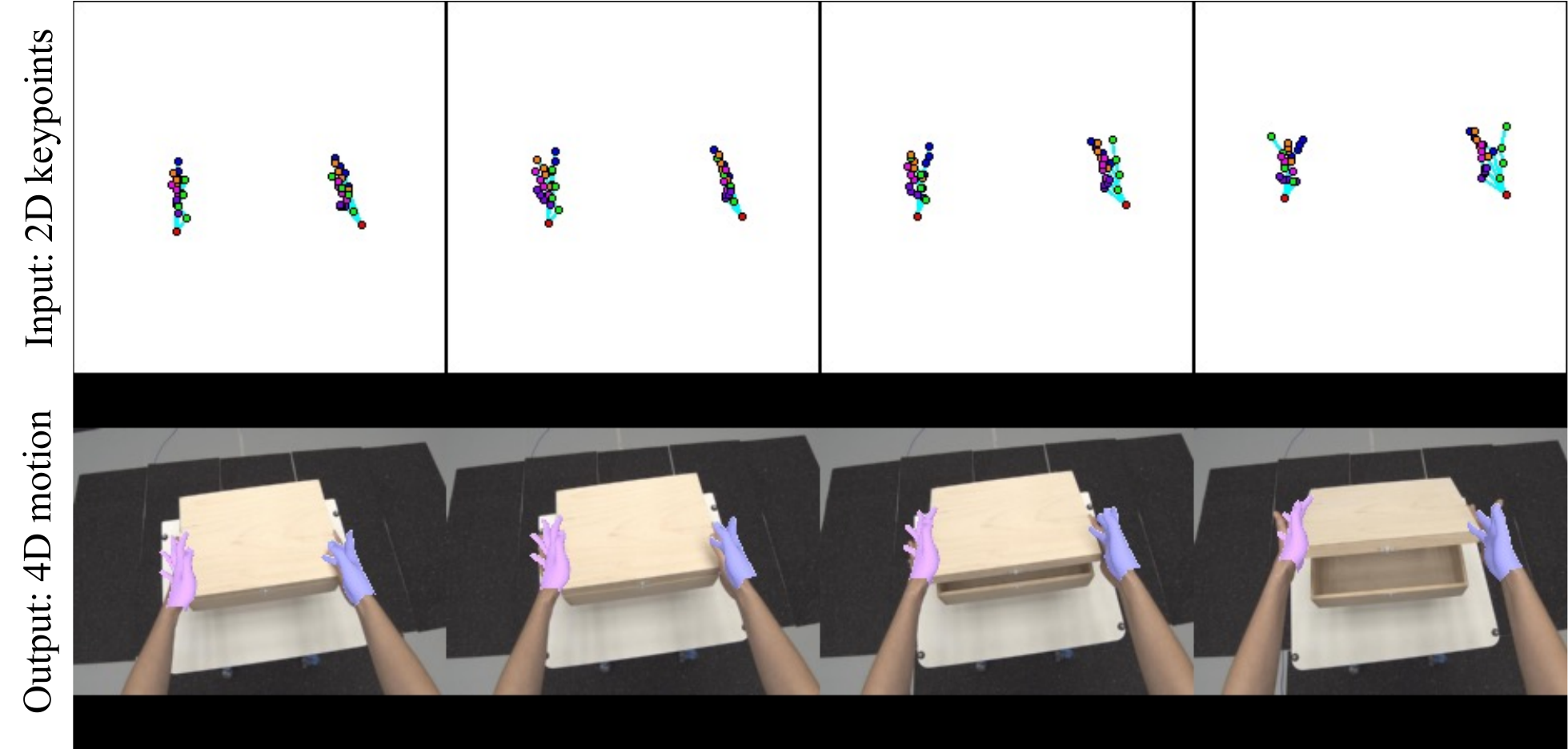}
    \hfill
    \includegraphics[width=0.495\textwidth]{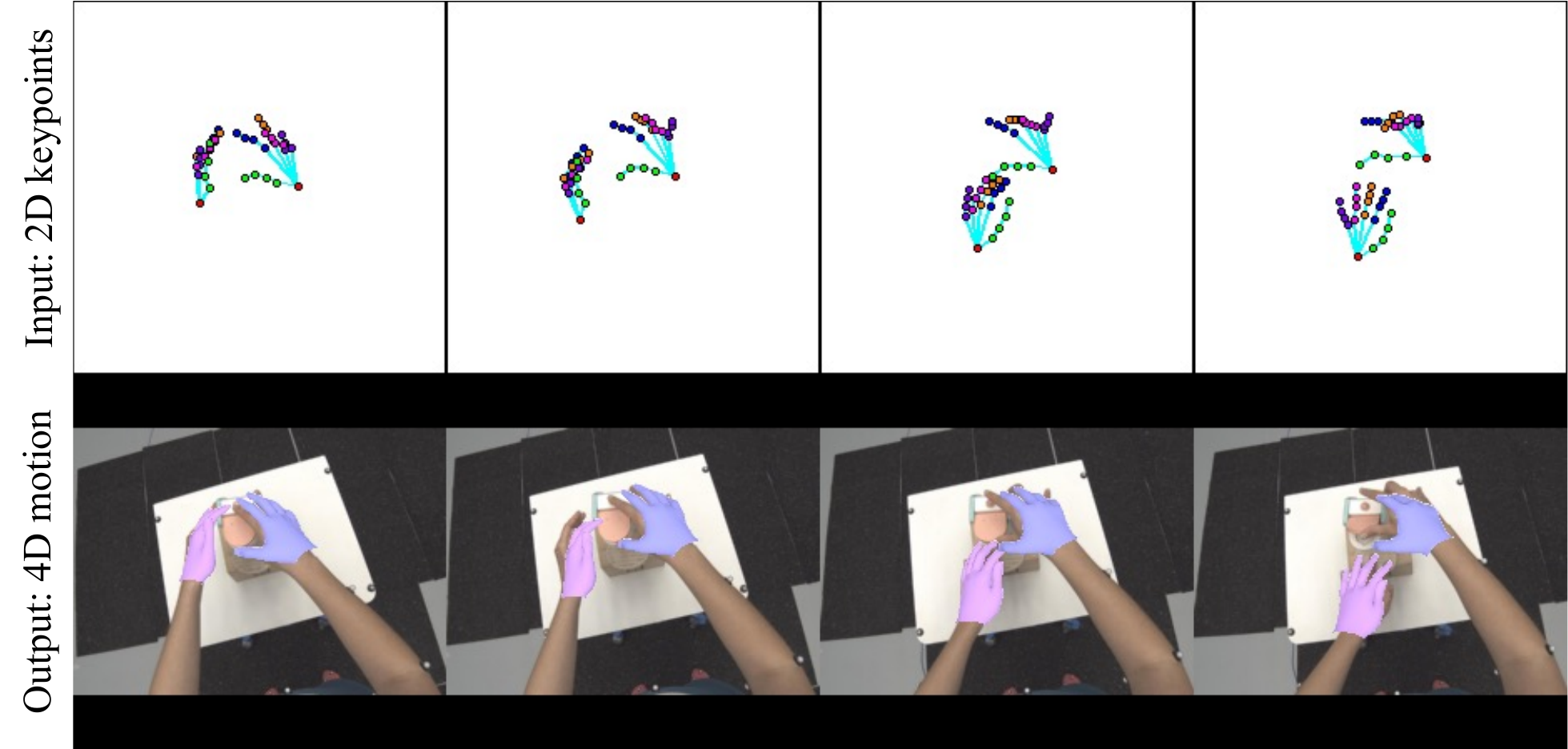}\\
    \vspace{0.1cm} 
    \hrule 
    \vspace{0.1cm}
    \includegraphics[width=0.495\textwidth]{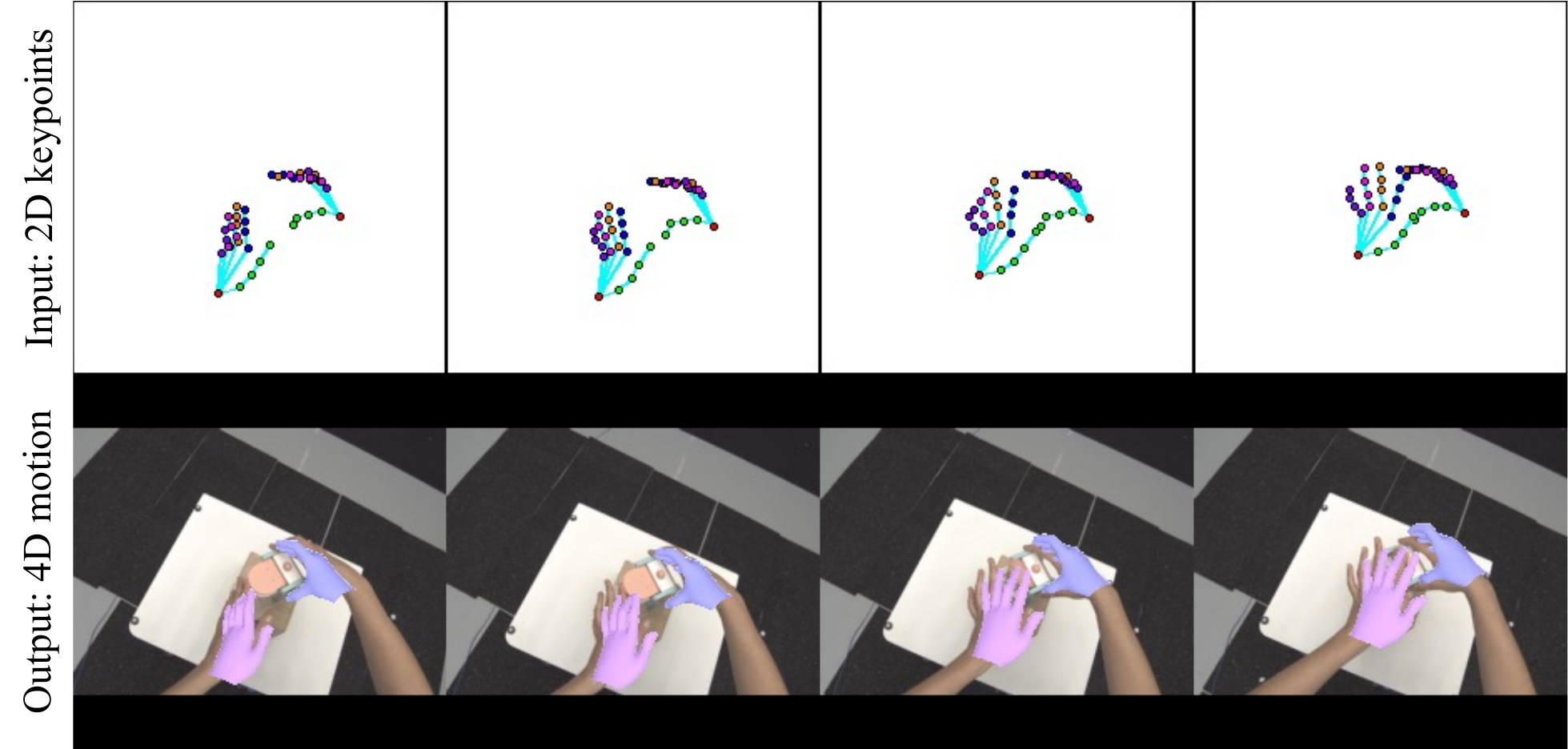}
    \hfill 
    \includegraphics[width=0.495\textwidth]{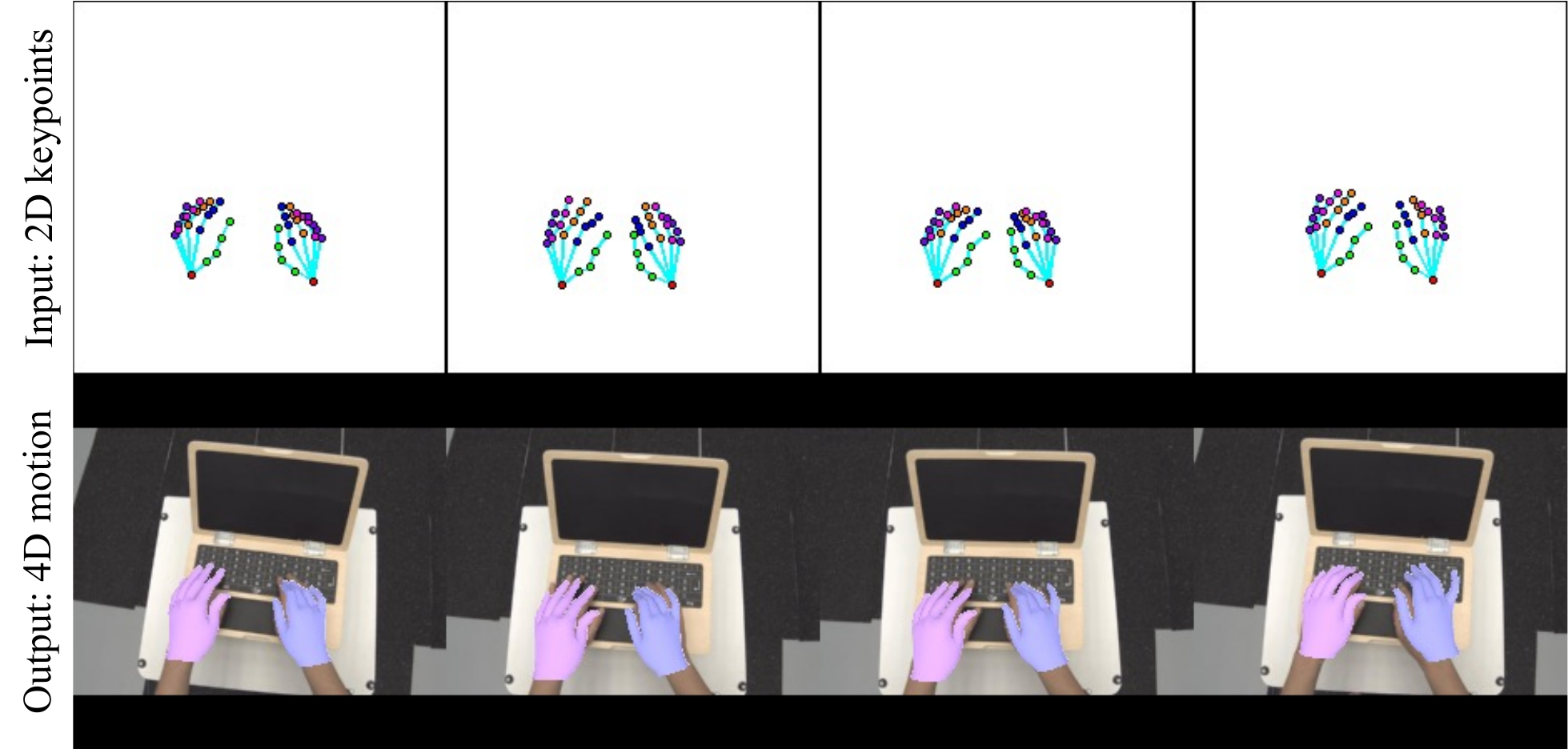}\\
    \vspace{0.1cm} 
    \hrule 
    \vspace{0.1cm}
    \caption{\textbf{4D hand predictions from the lifting model,} that outputs 3D MANO parameters from 2D keypoints \& camera parameter inputs. We show 4 frames with the MANO mesh rendered onto the image for visualization (images are not used as input).}
    \label{fig:lifting}
    \vspace{-0.7cm}
\end{figure*}

\noindent \textbf{Datasets}.
We use 5 lab datasets: H2O~\cite{kwon2021h2o}, H2O-3D~\cite{Hampali2022CVPR}, ARCTIC~\cite{Fan2023CVPR} Ego, HOT3D~\cite{Banerjee2024ARXIV}, \& DexYCB~\cite{Chao2021CVPR} with complete
3D annotations (\ie MANO labels) but limited data diversity. For diverse images, we include HoloAssist~\cite{Wang2023ICCV} \& AssemblyHands~\cite{ohkawa2023assemblyhands} (\ie incomplete annotations). The 2D keypoints on HoloAssist are estimated using HaMeR~\cite{pavlakos2024reconstructing}. AssemblyHands contain 3D \& 2D keypoints but no MANO labels.
We evaluate in 3 settings: (a) In-domain datasets: held-out test splits from training datasets (generalization to novel instances), (b) AssemblyHands~\cite{ohkawa2023assemblyhands}: a held-out test split that is not used for imputing labels, (c) EgoExo4D~\cite{Grauman2023EgoExo4D}: zero-shot generalization (not used for training in any way). More details about these datasets are provided in~\cref{tab:datasets}.

\subsection{Lifting Results}
\label{sec:lifting_res}
We start by evaluating the lifting model \& quality of imputed labels
(\cref{tab:assembly}, \cref{tab:lifting_ablation}).
We evaluate using 3D keypoint labels on AssemblyHands (not used for training
the lifting model).

\noindent \textbf{Comparisons to existing pseudo-labeling approaches}. We
consider: (a) predictions from
HaMeR~\cite{pavlakos2024reconstructing}, a high-performing single-image 3D hand
pose estimator, (b) HaWoR~\cite{Zhang2025ARXIV}, a recent method that uses
a data-driven motion priors to reconstruct the 3D hand motions in the
world-frame using videos, (c) EasyMocap~\cite{easymocap} that optimizes
3D MANO to conform to given 2D keypoints with temporal smoothing \& pose
regularization. Since EasyMocap \& our method require 2D keypoints as input, we use 2D keypoint predictions from HaMeR (\vs ground truth 2D keypoints) for both. We also modify EasyMocap to use ground truth camera poses in global motion initialization.

\cref{tab:assembly} shows that our lifting model produces the most accurate 3D labels across all metrics. These can be refined further via our reprojection
loss to better conform to the input 2D keypoints, which improves performance
further (last row in \cref{tab:assembly}). \cref{tab:gt_vs_pred} shows that our method outperforms EasyMocap even when EasyMocap has access to ground truth 2D keypoints. This suggests that together
with a strong 2D keypoint predictor like HaMeR, our lifting model can serve as an alternative to existing annotation pipelines that use human labeling together with EasyMocap.

\begin{table}
\caption{Our Lifting model is better than other methods for estimating 3D MANO
labels from videos on Assembly. We compare to: 1) a single-frame
feed-forward method HaMeR~\cite{pavlakos2024reconstructing}, 2) a video-based
feed-forward method HaWoR~\cite{Zhang2025ARXIV}, 3) an optimization-based
method EasyMocap~\cite{easymocap}. The top-part of the table shows that our
proposed lifting model outperforms all other methods across all metrics. Our
output can be further improved via 2D refinement (last row), which leads to
further gains. Here, both EasyMocap \& our method make use of 2D
hand keypoints predicted by HaMeR.}
\label{tab:assembly}
\centering
\vspace{-0.2cm}
\resizebox{0.7\linewidth}{!}
{
\begin{tabular}{l cccc}
\toprule
\multirow{2}{*}{\bf Method} & \multicolumn{4}{c}{\bf Assembly Hands} \\
\cmidrule(lr){2-5} & \bf \ma & \bf \mb & \bf \mc & \bf \md \\
\midrule
HaMeR~\cite{pavlakos2024reconstructing} & 29.7     & 6.8     & 16.1     & 36.9 \\
HaWoR~\cite{Zhang2025ARXIV}             & 61.6     & 6.1     & 46.3     & 131.1 \\
EasyMocap~\cite{easymocap}              & 44.5     & 8.0     & 31.9     & 24.1 \\
(Ours) Lifting                          & \bf 13.1 & \bf 5.2 & \bf 11.1 & \bf 17.8 \\
\midrule
\it (Ours) Lifting + 2D Refinement      & \it \phz4.2  & \it \phz3.3 & \it \phz3.6  & \it \phz5.4 \\
\bottomrule
\end{tabular}
}
\vspace{-0.5cm}
\end{table}

\begin{table}
\caption{Our lifting model alone is better than EasyMocap, even when EasyMocap
has access to ground truth keypoints (row 1 \vs row 5). Our full method
(lifting + 2D refinement) only suffers a minor degradation in performance when
switching from ground truth to predicted 2D keypoints (row 6 \vs row 3).}
\label{tab:gt_vs_pred}
\renewcommand{\arraystretch}{1.1}
\setlength{\tabcolsep}{4pt}
\centering
\vspace{-0.2cm}
\resizebox{0.8\linewidth}{!}
{
\begin{tabular}{l cccc}
\toprule
\multirow{2}{*}{\bf Method} & \multicolumn{4}{c}{\bf Assembly Hands} \\
\cmidrule(lr){2-5} & \bf \ma & \bf \mb & \bf \mc & \bf \md \\
\midrule
\multicolumn{5}{l}{\bf Using ground truth 2D keypoints} \\
\quad EasyMocap~\cite{easymocap}  & 36.3        & 7.7     & 16.8     & 15.4 \\
\quad (Ours) Lifting              & \phz9.6     & 5.1     & 10.0     & 11.7 \\
\quad (Ours) Lifting + 2D refinement & \bf \phz3.2  & \bf 2.7  & \bf \phz2.5 & \bf \phz3.7 \\
\midrule
\multicolumn{5}{l}{\bf Using 2D keypoints predicted from HaMeR~\cite{pavlakos2024reconstructing}} \\
\quad EasyMocap~\cite{easymocap}  & 44.5        & 8.0     & 31.9     & 24.1 \\
\quad (Ours) Lifting              &  13.1        & 5.2     & 11.1     & 17.8 \\
\quad (Ours) Lifting + 2D refinement & \bf \phz4.2  & \bf 3.3  & \bf \phz3.6 & \bf \phz5.4 \\
\bottomrule
\end{tabular}
}
\vspace{-0.7cm}
\end{table}

\begin{table}
\caption{{\bf Camera inputs for Lifting model} (\cref{sec:lifting}). Different ways of encoding camera parameters help, with Pl\"ucker rays being the most effective (this ablation uses ground truth 2D keypoints).}
\label{tab:lifting_ablation}
\centering
\vspace{-0.2cm}
\resizebox{0.8\linewidth}{!}
{
\begin{tabular}{l cccc cccc}
\toprule
\multirow{2}{*}{\bf Method} & \multicolumn{4}{c}{\bf Assembly Hands} \\
\cmidrule(lr){2-5} & \bf \ma & \bf \mb & \bf \mc & \bf \md \\
\midrule
No camera poses                        & 26.3        & 8.2     & 16.1     & 15.5 \\
Extrinsics + KPE                       & 17.9        & 7.6     & 16.4     & 15.8 \\
Pl\"ucker rays                         & 14.0        & 5.5     & 12.5     & 12.6 \\
Extrinsics + Pl\"ucker + KPE~~~~~~~ & \bf \phz9.6 & \bf 5.1 & \bf 10.0 & \bf 11.7 \\
\bottomrule
\end{tabular}
}
\vspace{-0.5cm}
\end{table}

\noindent \textbf{Ablations for lifting model.} In
\cref{tab:lifting_ablation} we see that conditioning only on 2D keypoints performs poorly. Injecting camera extrinsics (rotation, translation) \& intrinsics via angular encoding of 2D keypoints (KPE), in Row 2, helps quite a bit. Using Pl\"ucker rays in Row 3 also provides benefits. Our final model that uses all the different encodings together, performs the best. To maximally isolate the impact of these different strategies, these ablations use ground truth 2D keypoints (instead of predicted 2D keypoints as in \cref{tab:assembly} \& \cref{tab:gt_vs_pred}).

\noindent \textbf{Qualitative Visualizations.} \cref{fig:lifting} shows examples of 2D to 3D lifting achieved by our model. The lifting model accurately places and articulates the hands.

\begin{table*}
\vspace{-0.3cm}
\caption{\textbf{Comparisons to Methods based on the State-of-the-Art}. Our
\name model improves \ma \& \md by 36.02\% compared to static pose methods,
indicating significant hand movement in our testbed. It also outperforms
LatentAct~\cite{prakash2025how}, a recent VAE-based approach, adapted to our
setting, across all metrics.  Finally, \name is better than a
Transformer-based Regressor trained jointly on 3D \& 2D labels, where the 2D
labels are injected using a re-projection loss following SOTA
practices.
}
\label{tab:transformer}
\vspace{-0.3cm}
\centering
\setlength{\tabcolsep}{3pt}
\resizebox{0.99\linewidth}{!}
{
\begin{tabular}{l | gggg | cccc | cccc}
\toprule
\multirow{2}{*}{\bf Method} & \multicolumn{4}{g}{\bf In-domain datasets} & \multicolumn{4}{c}{\bf AssemblyHands} & \multicolumn{4}{c}{\bf EgoExo4D (Zero-shot)} \\
\cmidrule(lr){2-5} \cmidrule(lr){6-9} \cmidrule(lr){10-13}
       & \bf \ma & \bf \mb & \bf \mc & \bf \md & \bf \ma & \bf \mb & \bf \mc & \bf \md & \bf \ma & \bf \mb & \bf \mc & \bf \md \\
\midrule
Static Pose (trained in our setting)                       & 23.3     & 8.4     & 15.4     & 16.8     & 29.8     & 8.9     & 16.1     & 26.2     & 28.8     & 13.5     & 19.2     & 18.9 \\
Static Pose (from HaMeR~\cite{pavlakos2024reconstructing}) & 26.8     & 8.5     & 15.5     & 18.5     & 31.9     & 9.0     & \bf 14.2 & 38.8     & 32.7     & 13.6     & \bf 18.3 & 29.8 \\
LatentAct~\cite{prakash2025how} (adapted for our task)     & 17.7     & 7.4     & 16.2     & 17.6     & 21.3     & 9.1     & 17.2     & 22.8     & 26.4     & 13.5     & 19.5     & 49.9 \\
Transformer Regressor (3D + 2D sup.)                       & \bf 15.1 & 6.9     & 14.6     & 14.9     & 27.5     & 8.5     & 18.4     & 19.3     & 28.9     & \bf 13.0 & 21.3     & 19.0 \\
(Ours) \name                                               & 17.1     & \bf 6.5 & \bf 15.3 & \bf 13.5 & \bf 20.3 & \bf 8.4 & 15.4     & \bf 16.5 & \bf 18.8 & 13.2     & 18.9     & \bf 13.5 \\
\bottomrule
\end{tabular}}
\vspace{-0.5cm}
\end{table*}

\subsection{Forecasting Results}

Since there is no prior work that tackles this problem, we adapt recent work LatentAct~\cite{prakash2025how} to work in our setting and construct several baselines:\\
\textbullet\,\textbf{Static Pose Baseline} assumes a stationary hand \& uses 3D hand pose estimates on the input image as the forecast. We consider 2 variants: training a pose predictor in our setting and using outputs from off-the-shelf HaMeR~\cite{pavlakos2024reconstructing} (a high performing model trained on 10 datasets). \\
\textbullet\,\textbf{Transformer Regressor} uses the same architecture as our
model but directly regresses the future hand motion \& articulation. We consider
3 variants. {\bf 3D sup.} is trained with the same 3D ground truth as \name.
{\bf 3D sup. + 2D sup.} also uses 2D supervision via a reprojection loss on the
predicted 3D (following~\cite{pavlakos2024reconstructing,Prakash2024Handsb}).
This is only possible because the model directly regresses the 3D output. We
jointly train on 5 datasets with 3D labels \& 2 datasets with 2D labels. 
Finally, {\bf 3D sup. + 2D sup. + Imputed 3D.} uses additional 3D imputed
labels on top of the previous {\it 3D sup. + 2D sup.}
\\
\textbullet\, {\bf LatentAct~\cite{prakash2025how}} takes an image, text, contact point \& an interaction codebook (represented as the latent space of a VQVAE) as input to predict future 3D hand \& contact trajectory for a single hand. We adapt LatentAct to take only a single image as input and retrain it in our setting. \\
\textbullet\,\textbf{(Ours) \name } We consider 2 variants of our \name model, based on the supervision used. {\bf 3D sup.} is only trained on 5 datasets with 3D labels. {\bf 3D sup. + 2D sup.} is our final model that is trained jointly on 5 datasets with 3D labels \& the imputed  3D labels from our lifting model. \\
\textbullet\,We compare to other ways of using 2D labels with a diffusion model.
This amounts to attaching another head to make 2D predictions so that the image
backbone also gets gradients from 2D labels. We consider: \textbf{2D
Regression Head} \& \textbf{2D Diffusion Head} (denoising is done for 2D
keypoints).

\begin{table}
\vspace{-0.3cm}
\caption{\textbf{How to best incorporate 2D labels when training a diffusion model?} Labels from lifting model (Row 4) generalize better than alternatives based on attaching auxiliary 2D forecasting heads either diffusion (Row 3) or regression (Row 2).}
\label{tab:other}
\vspace{-0.2cm}
\centering
\setlength{\tabcolsep}{3pt}
\resizebox{0.99\linewidth}{!}
{
\begin{tabular}{l | gggg |cccc| cccc}
\toprule
\multirow{2}{*}{\bf Method} & \multicolumn{4}{g}{\bf In-domain datasets} & \multicolumn{4}{c}{\bf AssemblyHands} & \multicolumn{4}{c}{\bf EgoExo4D (Zero-shot)} \\
\cmidrule(lr){2-5} \cmidrule(lr){6-9} \cmidrule(lr){10-13}
       & \bf \ma & \bf \mb & \bf \mc & \bf \md & \bf \ma & \bf \mb & \bf \mc & \bf \md & \bf \ma & \bf \mb & \bf \mc & \bf \md \\
\midrule
No Additional 2D Supervision             & \bf 14.8 & \bf 5.9 & \bf 13.3 & \bf 12.4 & 27.9     & \bf 8.4 & 18.1     & 18.1     & 24.0     & \bf 13.0 & 20.9     & 19.6 \\
Inject 2D Sup. via a 2D Regression Head  & 15.5     & 6.4     & 14.0     & 13.8     & 26.8     & 8.5     & 16.6     & 16.9     & 25.9     & \bf 13.0 & 21.2     & 16.6 \\
Inject 2D Sup. via a 2D Diffusion Head   & 16.2     & 6.1     & 13.8     & 15.0     & 31.9     & \bf 8.4 & 18.2     & 19.8     & 24.8     & 13.2     & \bf 18.3 & 16.6 \\
(Ours) Inject 2D Sup. via Imputed Labels & 17.1     & 6.5     & 15.3     & 13.5     & \bf 20.3 & \bf 8.4 & \bf 15.4 & \bf 16.5 & \bf 18.8 & 13.2     & 18.9     & \bf 13.5 \\
\bottomrule
\end{tabular}
}
\vspace{-0.5cm}
\end{table}

\noindent \textbf{Static pose results}. The large values of \ma \& \md for static pose methods indicate that there is indeed a significant hand motion across timesteps since \ma \& \md are translation-focused metrics. \name leads to gains of 36.02\% on \ma \& \md across all settings. HaMeR scores the highest on \mc in EgoExo4D/Assembly, likely because it is trained across diverse images from 10 datasets.

\noindent{\bf Comparison with LatentAct}. 
\name outperforms the recent VQVAE-based LatentAct~\cite{prakash2025how} (adapted to our task) across all metrics. We see benefits of 16.4\% using our
\name model with \md gaining the most. This is likely due to LatentAct
requiring additional inputs, \ie, contact points \& text, to better
place the predicted motion in 3D space, which are not available in our setting.

\noindent \textbf{Peformance \vs Transformer Regressor.}
\name also outperforms on 11/12 metrics, a Transformer-based regression approach
that is trained using current SOTA practice of jointly training on 3D and 2D
labels, where the 2D labels are incorporated via a re-projection loss.  Gains
are particularly large in \ma \& \md in the zero-shot setting.

\noindent \textbf{Injecting 2D supervision improves performance on novel
datasets for \name.} Comparing Rows 1 \& 4 in \cref{tab:other}, we see that 2D supervision leads to large improvement in metrics on Assembly
\& EgoExo4D. Notably, \ma, \mc, \& \md improve by 10 -- 30\%. 
We also find that injecting 2D supervision mildly hurts performance on
in-domain datasets (Row 1 vs Row 4) for both models. We believe
this is because the same model now has to learn a much broader distribution
than what is tested in the in-domain datasets: \name may also suffer because the
imputed labels are not perfect (\cref{tab:assembly}). Nevertheless, injecting
2D labels helps by a lot on datasets without complete 3D annotations (\eg Assembly, EgoExo4D). We did not see a clear benefit by injecting imputed 3D labels on top of the 2D reprojection loss in Transformer regressor.

\begin{table*}
\setlength{\tabcolsep}{2.5pt}
\caption{\textbf{Breakdown of translation and articulation metrics in forecasting}.
{\bf (Row 1):} Static GT Articulation at $t=0$ + GT wrist pose at $t=t$
{\bf (Row 2):} GT Articulation at $t=t$ + Static GT wrist pose at $t=0$
{\bf (Row 3):} Full static GT pose, \ie Static GT Articulation at $t=0$ + Static GT wrist pose at $t=0$
{\bf (Row 4):} Full static predicted pose (model trained on our datasets), \ie Static Predicted Articulation at $t=0$ + Static Predicted wrist pose at $t=0$.
{\bf (Row 5):} Same as Row 4, but predictions come from off-the-shelf HaMeR~\cite{pavlakos2024reconstructing}. These highlight that translation constitutes a significant part of the metrics \& EgoExo4D involves much more dexterous actions.
}
\label{tab:analysis}
\resizebox{\linewidth}{!}{
\begin{tabular}{l | cc | gggg | cccc | cccc}
\toprule
\multirow{2}{*}{\bf Method} & \multirow{2}{*}{\bf Articulation} & \multirow{2}{*}{\bf Wrist Pose}
& \multicolumn{4}{g}{\bf In-domain datasets} & \multicolumn{4}{c}{\bf AssemblyHands} & \multicolumn{4}{c}{\bf EgoExo4D (Zero-shot)} \\
\cmidrule(lr){4-7} \cmidrule(lr){8-11} \cmidrule(lr){12-15}
       & & & \bf \ma & \bf \mb & \bf \mc & \bf \md & \bf \ma & \bf \mb & \bf \mc & \bf \md & \bf \ma & \bf \mb & \bf \mc & \bf \md \\
\midrule
                                                            & GT at $t=0$   & GT at $t=t$   & 4.5  & 3.0  & 4.6  & 2.3  & 5.0  & 3.6  & 5.9  & 3.4      & 15.1 & 10.5    & 25.1 & 2.6 \\
                                                            & GT at $t=t$   & GT at $t=0$   & 13.7 & 6.7  & 13.6 & 12.0 & 13.4 & 7.4  & 13.3 & 15.5     & 23.0 & 13.3    & 18.5 & 19.1\\
                                                            & GT at $t = 0$ & GT at $t=0$   & 15.4 & 8.2  & 15.2 & 12.0 & 14.1 & 8.6  & 13.9 & 15.5     & 22.7 & 13.5    & 18.2 & 19.1 \\
\midrule
Predictor (trained on our datasets)                           & Pred at $t=0$ & Pred at $t=0$ & 23.3 & 8.4  & 15.4 & 16.8 & 29.8 & 8.9  & 16.1 & 26.2     & 28.8 & 13.5    & 19.2 & 18.9 \\
Predictions from HaMeR~\cite{pavlakos2024reconstructing}    & Pred at $t=0$ & Pred at $t=0$ & 26.8 & 8.5  & 15.5 & 18.5 & 31.9 & 9.0  & 14.2 & 38.8     & 32.7 & 13.6    & 18.3 & 29.8 \\
\bottomrule
\end{tabular}}
\vspace{-0.5cm}
\end{table*}

\noindent \textbf{Proposed lifting \vs alternatives.}
In \cref{tab:other}, for diffusion model, the gains are marginal when injecting supervision via a 2D regression head or 2D diffusion head, obvious ways that are typically used with non-diffusion based models. However, imputing labels via our lifting model greatly improves \ma \& \md.

\noindent \textbf{Analysis of forecasting metrics}.
In~\cref{tab:analysis}, we analyze the impact of translation \& articulation on the metrics by using ground truth (GT) articulation \& wrist poses in different ways: {\bf Row 1}: This measures how much the hand articulation changes over the motion. {\bf Row 2}: This measures how much the wrist translates with respect to the first time step.
{\bf Row 3}: This considers changes in both articulation and translation as the motion progresses. 
{\bf Row 4}: Pose predictions for the given frame, copied over as the forecast for all future frames. Here predictions are coming from a model hand pose predictor trained on our datasets.
{\bf Row 5}: Same as Row 4, but predictions come from off-the-shelf HaMeR~\cite{pavlakos2024reconstructing}.

This analysis highlights that translation constitutes a significant part of the metrics and EgoExo4D involves much more dexterous actions compared to lab datasets. For evaluations involving GT poses, the \ma and \mc values should ideally be the same (as is the case with in-domain lab datasets) since the pose at first timestep is the same. However, that is not the case with AssemblyHands and EgoExo4D since they often contain invalid or missing labels for several joints due to which SVD does not converge during the procrustus alignment.

\begin{figure*}[t]
    \centering
    \setlength{\tabcolsep}{0pt}
    \begin{tabular}{c}
        \toprule
        \includegraphics[width=\textwidth]{gfx/viz_aria/header.pdf} \\ [-0.5em]
        \midrule 
        \includegraphics[width=\textwidth]{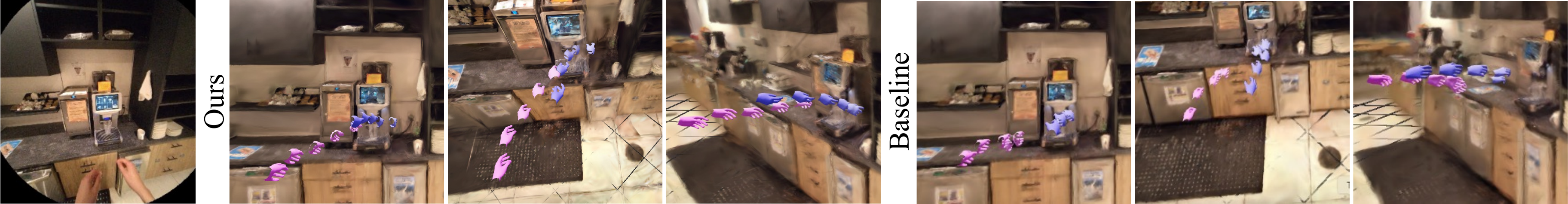} \\ [-0.25em]
        \midrule
        \includegraphics[width=\textwidth]{gfx/viz_aria/printer.pdf} \\ [-0.25em]
        \midrule
        \includegraphics[width=\textwidth]{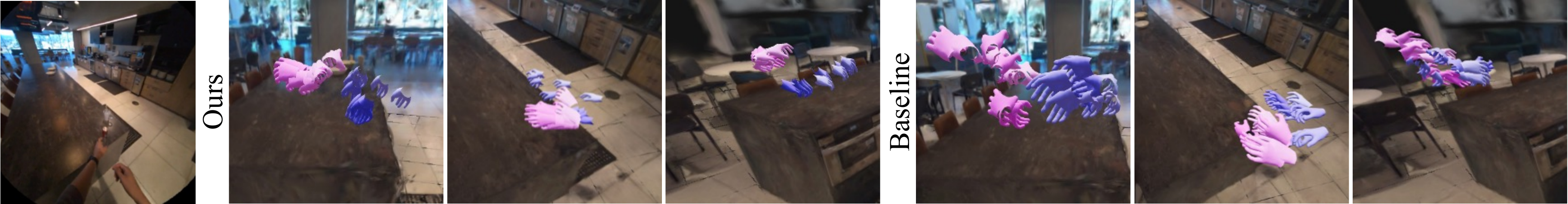} \\ [-0.25em]
        \bottomrule
    \end{tabular}
    \vspace{-0.2cm}
    \caption{\textbf{Qualitative comparison: \name vs. Transformer Regressor (3D + 2D sup.)} {\it Zero-shot} forecasts on unseen Aria images from~\cite{Yi2025CVPR}. Our predictions span longer trajectories, are smoother, better placed in the scene \& more plausible compared to the best baseline. The predicted motions highlight object interactions, \eg reaching for coffeemachine or the table. Left hand is shown in {\color{magenta}pink}, right hand in {\color{blue}blue}. Color saturation decreases as time proceeds, \ie further away timesteps in future are in lighter shades. We show the predicted motion in the 3D scene from 3 different views (3D scene is only for visualization \& these images are not seen during training).}
    \label{fig:viz}
    \vspace{-0.3cm}
\end{figure*}

\begin{figure}
    \centering
    \setlength{\tabcolsep}{0pt}
    \begin{tabular}{c}
        \toprule
        \includegraphics[width=\textwidth]{gfx/viz_aria/header.pdf} \\ [-0.5em]
        \midrule 
        \includegraphics[width=\textwidth]{gfx/multimodal_aria/coffee.pdf} \\ [-0.25em]
        \midrule
        \includegraphics[width=\textwidth]{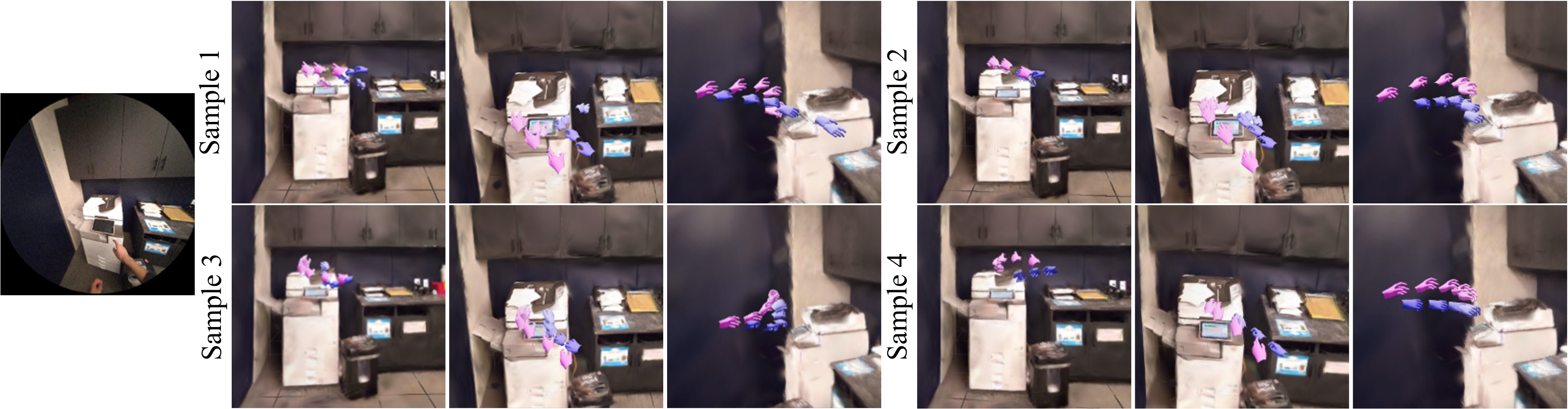} \\ [-0.25em]
        \midrule
        \includegraphics[width=\textwidth]{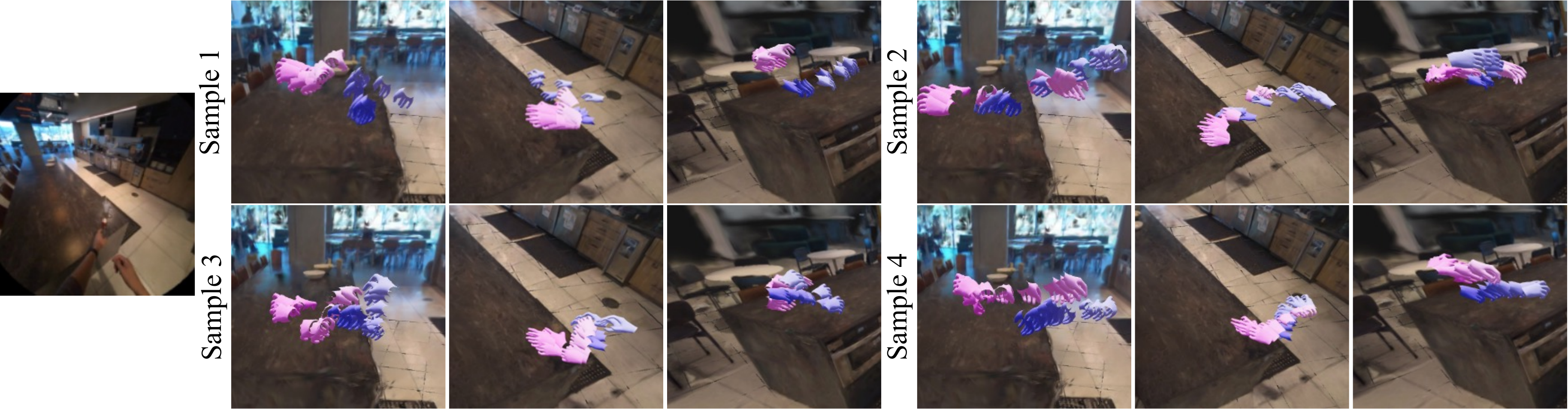} \\ [-0.25em]
        \bottomrule
    \end{tabular}
    \vspace{-0.3cm}
    \caption{\textbf{Different forecasts from the same input}.  We show 4 samples for 2 images from \name, indicating different modes of interaction like reaching \& grabbing towards different objects in the scene, \eg coffee machine, paper on the table, printer. We show the predicted motion in the 3D scene from 3 different views (3D scene is only for visualization \& these images are not seen during training).}
    \label{fig:multimodal}
    \vspace{-0.5cm}
\end{figure}

\noindent \textbf{Diversity \& multi-modality}. 
We compute these standard motion quality
metrics~\cite{guo2020action2motion,Tevet2023ICLR} on ARCTIC, which contains
ground truth MANO labels. The diversity
score for the ground truth distribution is 39.16. The predicted motion
distribution of \name (diversity = 41.14) is a lot closer to the ground
truth distribution compared to LatentAct (diversity = 302.45) \& transformer
regressor (diversity = 187.82). We also observe better mutimodality score for
our model (19.64 \vs 13.04 for LatentAct).
\cref{fig:multimodal} shows examples of multiple forecasts from the same input, \eg reaching \& grabbing motions towards different objects.

\begin{figure}
    \centering
    \begin{subfigure}{0.48\columnwidth}
        \centering
        \includegraphics[width=\linewidth]{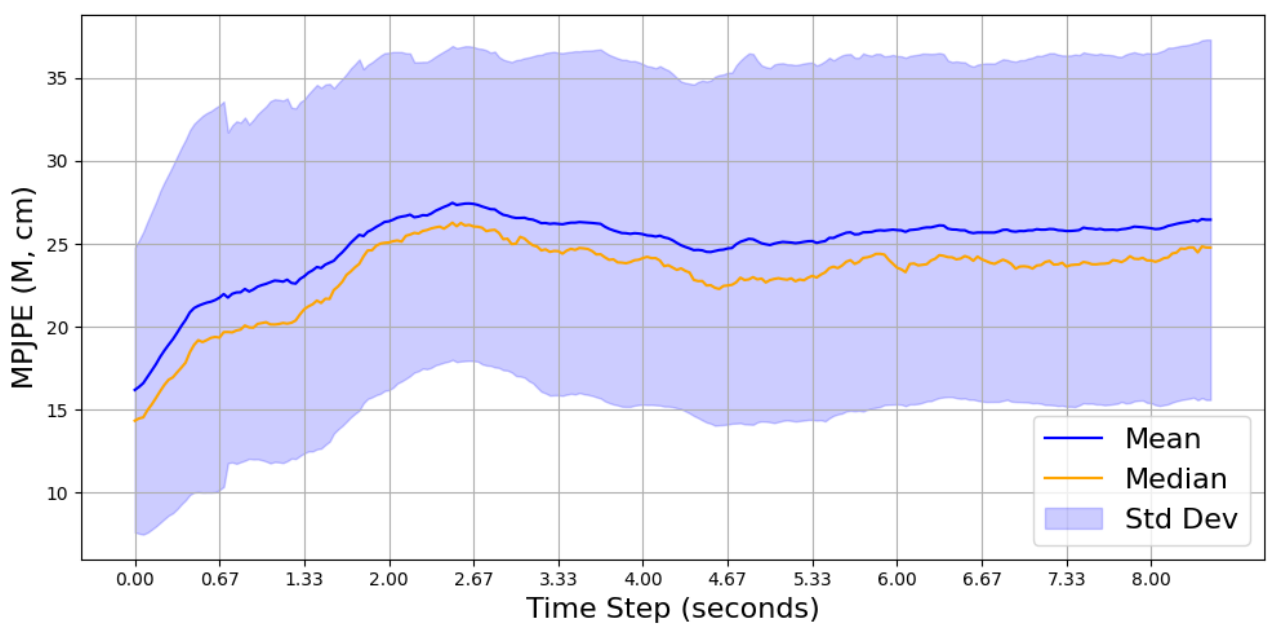}
    \end{subfigure}
    \hfill
    \begin{subfigure}{0.48\columnwidth}
        \centering
        \includegraphics[width=\linewidth]{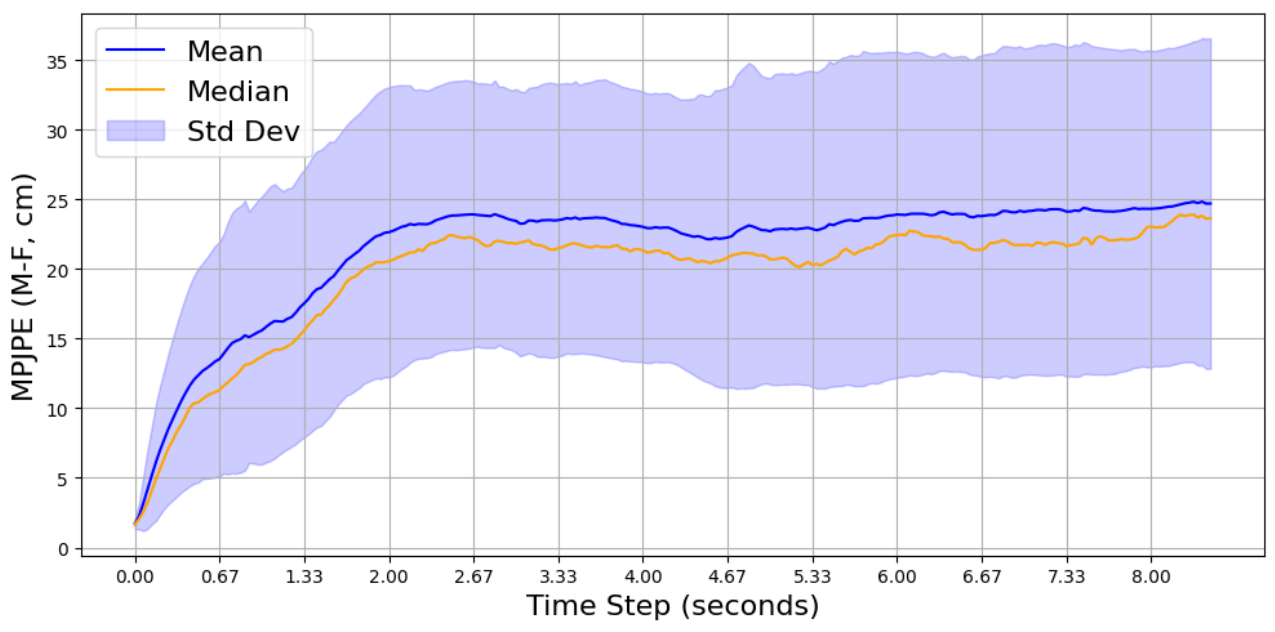}
    \end{subfigure}
    \vspace{-0.3cm}
    \caption{\textbf{Performance trends over time.} Forecasting gets harder for longer prediction horizons for both (top) \ma \& (bottom) \mc (Standard deviation is computed per time-step after aggregating errors from 5 generated motions for each sample).
    }
    \label{fig:time}
    \vspace{-0.5cm}
\end{figure}

\noindent \textbf{Performance trends over time}. In \cref{fig:time}, we see \ma (MPJPE) does not start from 0. This is because the model finds it hard to precisely predict the hand translation in the given frame (likely due to scale ambiguity in predicting metric 3D from a single image). \mc, where we factor out this imperfection by aligning to the ground truth hand in the first frame, shows a clear increasing trend.

\noindent\textbf{Qualitative comparisons.} We visualize the predicted motions for both our model \& Transformer Regressor (3D + 2D sup.) in \cref{fig:viz} on Aria images from~\cite{Yi2025CVPR}. Our motion predictions span longer trajectories, are smoother \& better placed in the scene compared to the baseline. The predicted motions also highlight object interactions, \eg reaching for coffeemachine or the table. See supplementary for more visualizations and analysis.

\noindent\textbf{Failure modes.} In egocentric videos, hands often go out of view as camera moves. This makes lifting hard since there is no input 2D keypoints for the model to use. For forecasting, heavy occlusions in the input image often leads to forecasted hand being incorrectly placed. As we are forecasting future motion, errors in the near timesteps compound over time (as seen in \cref{fig:time}).

\section{Conclusion}
We present a pipeline for lifting 2D hand keypoints in diverse videos to 3D MANO annotations and then use these labels for forecasting bimanual 3D hand motion from a single image in everyday settings. Our forecasting model consists of a conditional diffusion model trained with different types of supervision: 3D labels in lab datasets \& imputed 3D labels from diverse datasets using our lifting model. Our predictions span longer horizon, are smoother, better placed and capture multiple interaction modes, especially in zero-shot generalization settings. For future work, incorporating context , \eg past frames or intent and predicting object motion are important directions.

\boldparagraph{Acknowledgements} We thank Arjun Gupta, Shaowei Liu, Anand Bhattad \& Kashyap Chitta for feedback on the draft, and David Forsyth for useful discussion. This material is based upon work supported by NSF (IIS2007035), NASA (80NSSC21K1030), DARPA (Machine Common Sense program), Amazon Research Award, NVIDIA Academic Hardware Grant, and the NCSA Delta System (supported by NSF OCI 2005572 and the State of Illinois). Richard Li was supported by the NSF Institute for Artificial Intelligence and Fundamental Interactions (Grant No. PHY-2019786) and the Felicis Scholars program.

{
    \small
    \bibliographystyle{splncs04}
    \bibliography{biblioLong.bib, refs.bib}
}

\end{document}